\documentclass[onecolumn,british,twocolumn]{IEEEtran}
\usepackage[T1]{fontenc}
\usepackage[utf8]{inputenc}
\usepackage[active]{srcltx}
\usepackage{float}
\usepackage{amsmath}
\usepackage{amssymb}
\usepackage{graphicx}

\makeatletter

\floatstyle{ruled}
\newfloat{algorithm}{tbp}{loa}
\providecommand{\algorithmname}{Algorithm}
\floatname{algorithm}{\protect\algorithmname}

\usepackage{algorithm}
\usepackage{algpseudocode}
\usepackage{hyperref}

\makeatother

\usepackage{babel}
\begin{document}
\title{Compiling Deterministic Structure into SLM Harnesses}
\author{%
\begin{minipage}[t]{0.3\textwidth}%
\begin{center}
Zan-Kai Chong\\
\emph{School of Science and\\ Technology\\
Kwansei Gakuin University}\\
Japan\\
zankai@ieee.org
\par\end{center}%
\end{minipage}%
\begin{minipage}[t]{0.3\textwidth}%
\begin{center}
Hiroyuki Ohsaki\\
\emph{School of Science and\\ Technology\\
Kwansei Gakuin University}\\
Japan\\
ohsaki@kwansei.ac.jp
\par\end{center}%
\end{minipage}%
\begin{minipage}[t]{0.3\textwidth}%
\begin{center}
Bryan Ng\\
\emph{ 
School of Engineering \& Computer Science\\
Victoria University of Wellington}\\
New Zealand
\\
ckbryan@hotmail.com
\par\end{center}%
\end{minipage}}
\maketitle
\begin{abstract}
Enterprise deployment of small language models (SLMs) is constrained
by epistemic asymmetry: small models cannot reliably self-correct,
while frontier LLMs face prohibitive costs and data-sovereignty restrictions.
We propose Semantic Gradient Descent (SGDe), a teacher-student framework
that compiles agentic workflows offline into discrete execution plans
$\theta=\{\mathcal{G},\mathcal{P},\mathcal{C}\}$ comprising a DAG
topology, system prompts, and deterministic executable code. A frontier
teacher generates natural-language critiques that function as directional
gradients, iteratively rewriting the SLM's harness. Within a Probably
Approximately Correct (PAC) framework, we bound the effective hypothesis
space by $\text{O}(3^{k})$ in the number of subtask types $k$, enabling
convergence from as few as three training examples on structurally
homogeneous tasks. On a GSM-Hard-derived test set, compiled workflows
reach 91.3\% accuracy at $m=5$ and 99.3\% at $m=3$ (+26.3\% to +34.3\%
over the leading intra-substrate baseline, DSPy). SGDe lifts the offload/retain
decision of PAL and PoT from a static whole-problem template to a
per-node, trace-driven optimisation target, and combines it with a
structural-consensus mechanism that wraps variance-sensitive steps
in fan-out/fan-in subgraphs aggregated by deterministic voting.
\end{abstract}

\section{Introduction}\label{sec:Introduction}

Frontier large language models (LLMs) such as GPT-5, Claude, and Gemini
reason across diverse domains without task-specific fine-tuning, yet
in enterprise settings they are often impractical due to inference
cost, regulatory compliance, and data-sovereignty constraints. Organisations
therefore deploy specialised systems for narrow, well-defined document
types \cite{ling2024domainspecializationkeymake,lu2025smalllanguagemodelssurvey},
reminiscent of the expert systems of the 1970s \cite{brock2022expert}.
Small language models (SLMs, sub-billion to \textasciitilde 10B parameters
\cite{wang2024comprehensivesurveysmalllanguage}) are well suited
here, achieving comparable performance at a fraction of the cost,
but they suffer from \emph{epistemic asymmetry}: the same agent cannot
reliably perform a task and judge its own reasoning, so hallucinations
go undetected \cite{zhang2024smalllanguagemodelsneed,huang2024largelanguagemodelsselfcorrect}.

Existing approaches to SLM deployment fall short on the discrete structure
of the workflow. Knowledge distillation \cite{ballout2024efficientknowledgedistillationempowering}
transfers behaviour via continuous-weight updates, and trajectory-based
agentic distillation \cite{zeng2023agenttuningenablinggeneralizedagent}
extends this to action sequences but hides the asymmetry rather than
resolving it \cite{liu2025agentbenchevaluatingllmsagents}. A second
line of work treats prompts and topology as optimisable: TextGrad
\cite{yuksekgonul2024textgradautomaticdifferentiationtext} back-propagates
textual gradients, Wang et al. \cite{wang2024correctlysemanticbackpropagationlanguagebased}
formalise semantic backpropagation, APO \cite{pryzant2023automaticpromptoptimizationgradient}
and DSPy \cite{khattab2023dspycompilingdeclarativelanguage} refine
prompts against a fixed pipeline, and AFlow \cite{zhang2025aflow}
mutates the DAG via Monte Carlo Tree Search atop manually-engineered
topologies (AutoGen, Graph of Thoughts \cite{Besta_2024,wu2023autogenenablingnextgenllm}).
Across these frameworks every node remains an LLM call: the unit of
optimisation is text, connectivity, or both, but never the \emph{substrate},
i.e., the set of computations the LLM is responsible for. We call
these \emph{intra-substrate optimisers}. Program-aided reasoning (PAL
\cite{gao2023pal}, PoT \cite{chen2023pot}) does cross the substrate
boundary, but only by always prompting the LLM to emit a Python program
for the entire question. The offload decision is fixed in advance,
not chosen per node or per task.

We propose Semantic Gradient Descent (SGDe), a PAC-bounded approach
to harness engineering \cite{lopopolo2026harness,trivedy2026anatomy}
that compiles a fully executable harness $\theta=\{\mathcal{G},\mathcal{P},\mathcal{C}\}$
offline, where $\mathcal{G}$ is a DAG topology,  $\mathcal{P}$ a
set of specialised prompts, and $\mathcal{C}$ deterministic code.
SGDe expands the unit of optimisation to the substrate itself, lifting
the offload/retain decision from a fixed prompting convention to a
per-node, trace-driven optimisation target, and replaces continuous
parameter updates with a teacher-student loop in which a frontier
teacher analyses traces, attributes errors, and rewrites the workflow
artefacts. The distinction is load-bearing for SLM deployment: for
frontier LLMs the gap between a prompt and a Python function is narrow
and intra-substrate optimisation captures most gains, but for SLMs
the same gap is precisely where deployment fails. Substrate compilation
is composable with intra-substrate search, so we evaluate SGDe against
the leading intra-substrate baseline (DSPy) to isolate its contribution.

The contributions of this paper are threefold. First, we introduce
\emph{substrate compilation} as a new unit of optimisation: at every
node the teacher selects adaptively among prompt refinement, capability
offloading (rewriting a node as deterministic code), and structural
consensus (wrapping a node in a fan-out/fan-in ensemble with deterministic
aggregation), directly addressing the capability and reliability gaps
that motivate SLM deployment. Second, we derive a resource-bounded
PAC analysis in which the effective hypothesis space is bounded by
$|\Theta_{\mathcal{T}}|=\text{O}(3^{k})$ in the number of distinct
subtask types $k$, explaining mechanistically why SGDe converges
from as few as three training examples on structurally homogeneous
tasks. Third, we empirically validate SGDe on a GSM-Hard-derived test
set, achieving 91.3\% accuracy at $m=5$ and 99.3\% at $m=3$, a +26.3\%
to +34.3\% absolute improvement over the leading intra-substrate baseline
(DSPy).

\section{Semantic Gradient Descent (SGDe) Architecture}\label{sec:Semantic-Gradient-Descent}

This section formalises the SGDe architecture in detail, anchored
in a running example that is threaded through the subsequent subsections.

\subsection{Running Example}\label{subsec:Running-Example}

To anchor the framework in a concrete task and reduce the cognitive
load of introducing the architectural components in parallel, we use
a single GSM-Hard task as a running example throughout this section,
and revisit it in Section \ref{subsec:Experiment-2:-Sample}:

\emph{For his birthday, Jamal received a cash gift sufficient to buy
one 55-dollar video game and one 25-dollar controller with 10 dollars
left over. How many dollars did Jamal receive?} (gold answer: 90).

Solving this requires parsing the item prices and the leftover amount,
then applying the rule “amount received = total cost + leftover” ($90=55+25+10$).
The narrative is mildly inverted relative to the more common “change
received” framing: Jamal \emph{receives} the gift rather than paying,
a property we exploit in Section \ref{subsec:Experiment-2:-Sample}
to differentiate the $m=3$ and $m=5$ compiled harnesses.

At $t=0$, the workflow is a single monolithic chain-of-thought node
that reads the question, reasons through cost and leftover, and emits
the final number. The 1.5B-parameter student frequently parses the
operands correctly but fails at arithmetic, prompting the $t=1$ update,
namely the capability-offloading refinement of Section \ref{subsec:The-Capability-Offloading}
(Figure \ref{fig:offloading}), which decomposes the monolithic node
into an LLM extraction node followed by a deterministic Python evaluation
node. Subsequent iterations may further wrap the extraction step in
a structural-consensus subgraph (Section \ref{subsec:fan-in-fan-out},
Figure \ref{fig:Structural-consensus}) when the teacher detects high
variance. The fully compiled harness produced after termination is
the object evaluated in Section \ref{subsec:Experiment-2:-Sample}.

\subsection{Two Levels of Optimisation: Granularity and Compilation}\label{subsec:three-actions}

SGDe's optimiser operates at two levels. Granularity operations (decomposition,
fusion; see Section \ref{subsec:Granularity-Adjustment-via}) reshape
the subtask partition without altering any node's substrate. Compilation
actions then select at each $v\in V$ among three mutually exclusive
substrate choices: (i) prompt refinement (revised $p\in\mathcal{P}$);
(ii) capability offloading (Section \ref{subsec:The-Capability-Offloading}),
re-typing $v$ as a code node $c\in\mathcal{C}$; and (iii) structural
consensus (Section \ref{subsec:fan-in-fan-out}), replicating $v$
under perturbed prompts with deterministic vote aggregation. The three
actions address orthogonal failure modes (instruction quality, capability
gap, output variance). The per-node choice is driven by the teacher's
semantic gradient $g_{\text{sem}}$ rather than fixed in advance.
This two-level, adaptive, per-node unification distinguishes SGDe
from prior work in which each action is applied in isolation or globally.
The iterative lifecycle in which these actions are applied (execution,
attribution, candidate generation, and greedy acceptance) is summarised
in Algorithm \ref{alg:Semantic-gradient-descent}. Subsequent subsections
detail each action.

\subsection{The Capability Offloading Principle}\label{subsec:The-Capability-Offloading}

PAL \cite{gao2023pal} and PoT \cite{chen2023pot} established that
deterministic sub-computations are better delegated to a Python interpreter
than to an LLM's token generator, but as a static whole-problem pattern.
SGDe lifts this to a per-node, trace-driven decision: at each $v\in V$
the teacher evaluates, from empirical failure attribution, whether
$v$ is reliably executable by code, and if so rewrites it as $c\in\mathcal{C}$.
The execution plan $\theta=\{\mathcal{G},\mathcal{P},\mathcal{C}\}$
is therefore a partition of the task between the probabilistic student
and the deterministic runtime, and it is this partition that SGDe
optimises.

\begin{figure}
\begin{centering}
\includegraphics[width=0.8\columnwidth]{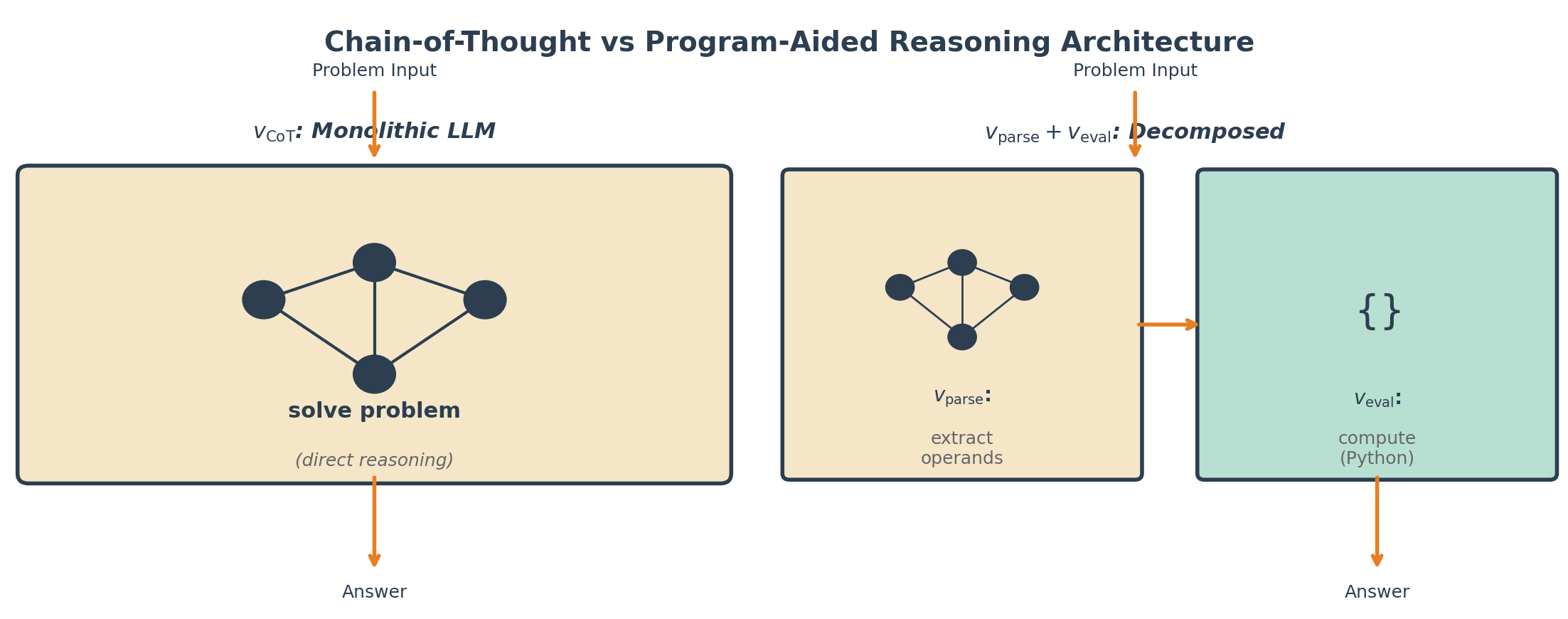}
\par\end{centering}
\caption{The capability offloading principle, illustrated on a compound teacher
update. A monolithic reasoning node $v_{\text{CoT}}$ (left) is first
decomposed into a parsing sub-node $v_{\text{parse}}$ and an evaluation
sub-node $v_{\text{eval}}$ (a granularity operation on $\mathcal{G}$),
after which $v_{\text{eval}}$ is offloaded from the LLM substrate
to deterministic Python in $\mathcal{C}_{t}$ (right, a compilation
action). The SLM is restricted to operand extraction. Arithmetic is
delegated to code, eliminating the epistemic-asymmetry failure mode
on numerically intensive tasks.}\label{fig:offloading}
\end{figure}

Figure \ref{fig:offloading} illustrates the principle on the running
example of Section \ref{subsec:Running-Example}. The initial monolithic
chain-of-thought node $v_{\text{CoT}}$ is rewritten by one SGDe iteration
into an LLM extraction node $v_{\text{parse}}\in\mathcal{P}$ (emitting
a structured tuple of operands and operator) followed by a deterministic
Python node $v_{\text{eval}}\in\mathcal{C}$. The student's role at
$v_{\text{eval}}$ is compiled away entirely. Capability offloading
and the complementary structural-consensus mechanism (Section \ref{subsec:fan-in-fan-out}),
which wraps high-variance reasoning steps rather than replacing them,
jointly produce the gains reported in Section \ref{sec:Experiment-Evaluation}.
Both are present in every winning harness.

\subsection{Granularity Adjustment via Cognitive Load Balancing}\label{subsec:Granularity-Adjustment-via}

Granularity adjustment operates within the LLM substrate, resizing
the scope of nodes that remain LLM calls without changing their type.
The teacher analyses traces $\text{Tr}_{t}$ to identify cognitive
bottlenecks, i.e., nodes $v$ where the student fails due to context
bloat or attention degradation.
\begin{itemize}
\item \emph{Node decomposition} splits an overloaded $v$ into a chain $v_{1}\rightarrow\cdots\rightarrow v_{k}$
such that $\bigcup_{i}\text{scope}(v_{i})=\text{scope}(v)$, with
each sub-node receiving a narrower prompt $p_{i}\in\mathcal{P}_{t+1}$.
\item \emph{Node fusion} merges consecutive $v_{i},v_{i+1}$ when their
per-node accuracies on $\text{Tr}_{t}$ are statistically indistinguishable
and $\hat{R}(\theta_{\text{merge}})-\hat{R}(\theta_{t})<\epsilon_{\text{cost}}$,
reducing latency without accuracy loss.
\end{itemize}
Granularity operations determine the subtask partition. The three
compilation actions of Section \ref{subsec:three-actions} then determine
how each subtask is executed. A single teacher update may combine
both, as in Figure \ref{fig:offloading}.

\subsection{Structural Consensus Patterns (Fan-Out/Fan-In)}\label{subsec:fan-in-fan-out}

\begin{figure}
\begin{centering}
\includegraphics[width=0.8\columnwidth]{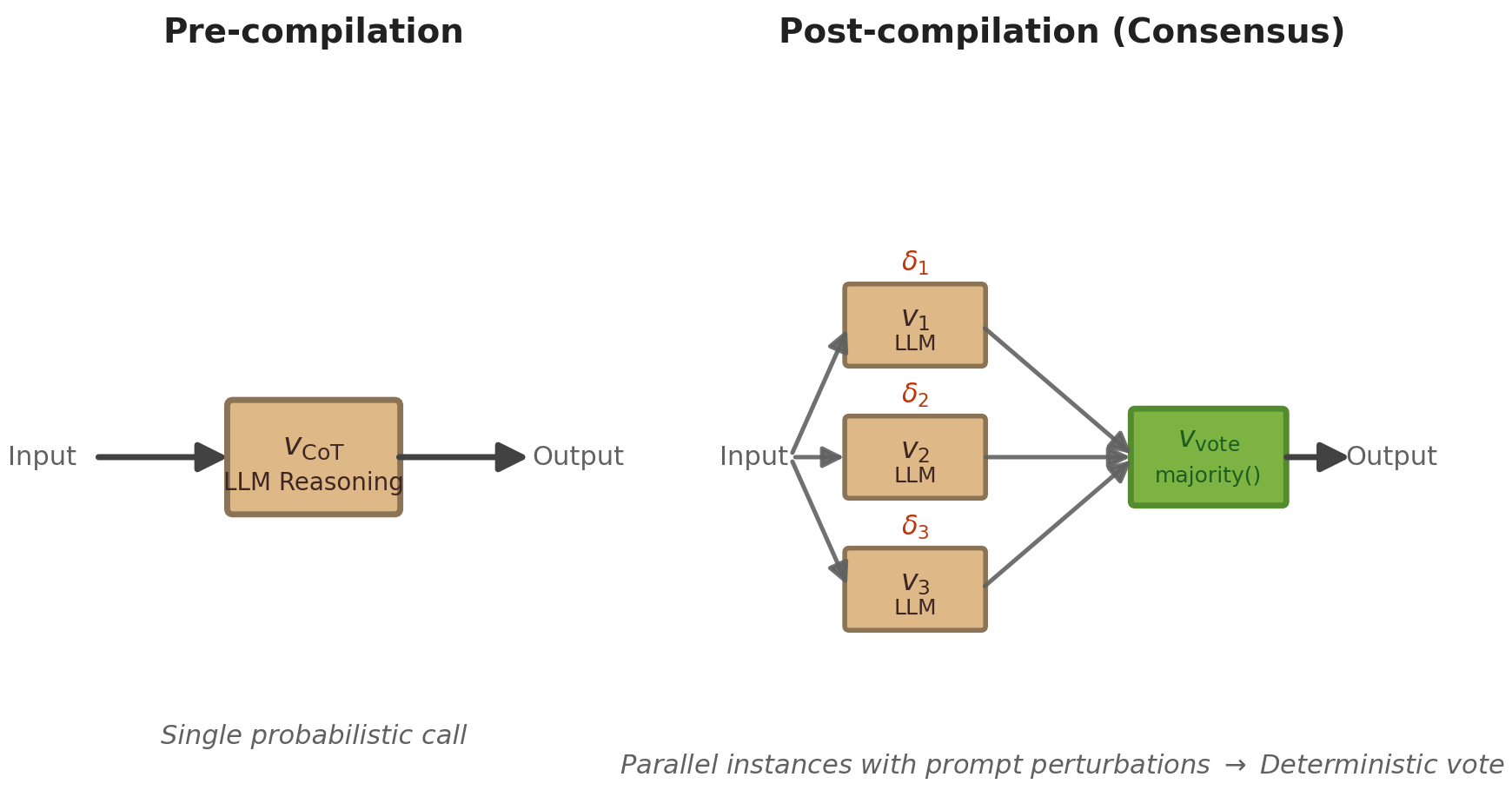}
\par\end{centering}
\caption{Structural consensus via fan-out/fan-in compilation. A single probabilistic
reasoning node $v$ (left) is replicated by the teacher into three
parallel LLM instances with prompt perturbations $\left\{ \pi_{1},\pi_{2},\pi_{3}\right\} $
and aggregated by a deterministic voting node $v_{\text{vote}}\in\mathcal{C}_{t}$
(right). Unlike capability offloading (Figure \ref{fig:offloading}),
the student's probabilistic inference is retained and replicated rather
than replaced. The deterministic structure is confined to the aggregation
step. }\label{fig:Structural-consensus}
\end{figure}

Structural consensus addresses the complementary case to capability
offloading: the student is capable but unreliable, and output variance
dominates the error. When $g_{\text{sem}}$ reports high variance
at a specific reasoning step, the teacher compiles a fan-out/fan-in
subgraph $\mathcal{G}_{\text{cons}}=(\{v_{\text{in}},v_{1},v_{2},v_{3},v_{\text{vote}}\},\cup_{i}\{(v_{\text{in}},v_{i}),(v_{i},v_{\text{vote}})\})$
directly into $\mathcal{G}$. Figure \ref{fig:Structural-consensus}
illustrates this transformation: the parallel instances $v_{1},v_{2},v_{3}$
carry perturbed prompts $\{p+\pi_{i}\}\subset\mathcal{P}_{t+1}$,
and $v_{\text{vote}}\in\mathcal{C}_{t+1}$ performs a deterministic
aggregation (e.g., majority vote). The teacher applies this pattern
only where the expected risk reduction $\Delta\hat{R}$ exceeds the
marginal compute cost.

\begin{algorithm}
\caption{Semantic gradient descent (SGDe) for agentic workflows. The teacher
iteratively critiques execution traces and rewrites the execution
plan $\theta=\left\{ \mathcal{G},\mathcal{P},\mathcal{C}\right\} $,
accepting candidates via greedy hill-climbing, until the empirical
risk falls below $\epsilon$ or the epoch budget $T$ is reached.
}\label{alg:Semantic-gradient-descent}

\begin{algorithmic}[1]
\Require 
Student agent $\mathcal{S}$; Teacher agent $\mathcal{T} = (\mathcal{T}_{\text{critic}}, \mathcal{T}_\text{optimiser})$;
Training sample $\mathcal{Z} =\{(\tau_i,y_i^*)\}_{i=1}^m\sim\mathcal{D}$;
Initial Execution Plan $\theta_0 =\{\mathcal{G}_0,\mathcal{P}_0,\mathcal{C}_0\}$;
Master Prompt $\Phi$;
Acceptable error $\epsilon$;
Max epochs $T$
\Ensure Optimised Execution Plan $\theta_{\text{opt}}$
\State Initialise: $\theta_{\text{best}} \leftarrow \theta_0$
\For{epoch $t = 0, 1, \ldots, T$}
    \State \Comment{Forward Pass: Constrained Execution}
    \State Execute Student on $\mathcal{Z}$: $\hat{y}_t,\text{Tr}_t\leftarrow\mathcal{S}(\mathcal{Z}; \theta_{\text{best}})$
    \State Calculate Risk: $\hat{R}_{\text{current}} \leftarrow \frac{1}{m}\sum_{i=1}^{m} \mathcal{L}(\hat{y}_{t,i}, y^*_i, \text{Tr}_{t,i})$
    \If{$\hat{R}_{\text{current}} \leq \epsilon$}
        \State \Return $\theta_{\text{opt}} \leftarrow \theta_{\text{best}}$
    \EndIf
    \State \Comment{Backward Pass: Semantic Error Attribution}
    \State Generate Critique: $g_{\text{sem}} \leftarrow \mathcal{T}_{\text{critic}}(\Phi, \text{Tr}_t, \hat{R}_{\text{current}})$
    \State \Comment{Greedy Candidate Generation}
    \State Generate Candidate: $\tilde{\theta} \leftarrow \mathcal{T}_{\text{optimiser}}(\Phi, g_{\text{sem}}, \theta_{\text{best}})$
    \State Execute Student on $\mathcal{Z}$ with $\tilde{\theta}$ to estimate $\hat{R}(\tilde{\theta})$
    \If{$\hat{R}(\tilde{\theta}) < \hat{R}_{\text{current}}$}
        \State $\theta_{\text{best}} \leftarrow \tilde{\theta}$ \Comment{Keep the improved DAG}
    \EndIf
\EndFor
\State \Return ``Optimisation Failed to Converge''
\end{algorithmic}
\end{algorithm}

\section{Mathematical Model }\label{sec:Mathematical-Model}

The semantic loss reduces to a scalar that triggers updates. The corrective
signal itself is the semantic gradient, the teacher's natural-language
verdict on the execution trace, used to rewrite a discrete artefact
rather than to take a metric step. We now formalise the parameterisation
and optimisation procedure.

\subsection{Forward Pass and Empirical Risk}

At iteration $t$, the execution plan $\theta_{t}=\{\mathcal{G}_{t},\mathcal{P}_{t},\mathcal{C}_{t}\}$
(cf. Section \ref{sec:Introduction}) introduces an iteration index,
with the DAG made explicit as $\mathcal{G}_{t}=(V,E)$, and the teacher
$\mathcal{T}$ guided by a master prompt $\Phi$ encoding the immutable
optimisation directives. To satisfy PAC learning requirements, optimisation
is performed over a batch of tasks rather than a single instance.
Let $\mathcal{D}$ be the real-world joint distribution of tasks and
their ground-truth answers within the target domain. We draw a finite
training set of $m$ paired samples, $\mathcal{Z}=\left\{ \left(\tau_{1},y^{*}_{1}\right),\cdots,\left(\tau_{m},y^{*}_{m}\right)\right\} \sim\mathcal{D}$.

For each $\tau_{i}$ the student emits $\hat{y}_{t,i}=\mathcal{S}(\tau_{i};\theta_{t})$
and a trace $\text{Tr}_{t,i}$. The composite semantic loss $\mathcal{L}(\hat{y},y^{*},\text{Tr})$
scores both task accuracy and inference cost.

The empirical risk 
\begin{equation}
\hat{R}\left(\theta_{t}\right)=\frac{1}{m}\sum^{m}_{i=1}\mathcal{L}\left(\mathcal{S}\left(\tau_{i};\theta_{t}\right),y^{*}_{i},\text{Tr}_{t,i}\right)
\end{equation}
 approximates the true generalisation error $R(\theta_{t})=\mathbb{E}_{\mathcal{D}}[\mathcal{L}]$.
Exceeding the threshold triggers the backward pass.

\subsection{Semantic Error Attribution (The Backward Pass)}

Because the parameter space of the execution plan $\theta_{t}$ is
entirely discrete, calculating a standard numerical gradient $\nabla_{\theta}\hat{R}$
is mathematically undefined. To overcome this, our framework replaces
the continuous backpropagation with an LLM-driven heuristic update
aimed at minimising the empirical risk $\hat{R}\left(\theta_{t}\right)$.

We therefore introduce a two-role teacher: the critic $\mathcal{T}_{\text{critic}}$
produces the semantic gradient $g_{\text{sem}}$ (a natural-language
critique that functions as a discrete directional signal), while the
separate optimiser $\mathcal{T}_{\text{optimiser}}$ consumes this
signal and emits a candidate successor plan via $\theta_{t+1}=\mathcal{T}_{\text{optimiser}}(\Phi,g_{\text{sem}},\theta_{t})$.
The optimiser accepts a candidate only if it reduces $\hat{R}$ (greedy
hill-climbing, Algorithm \ref{alg:Semantic-gradient-descent}).

\subsubsection{Fault Localisation (Semantic Gradient Formulation)}

When the empirical risk exceeds the acceptable threshold, the teacher
agent initiates the backward pass by evaluating the execution traces
($\text{Tr}_{t}$) of the training set. Acting as a diagnostic critic,
it performs semantic error attribution to generate a natural language
critique $g_{\text{sem}}$. This critique, which is the semantic gradient,
functions as a discrete directional gradient, isolating the exact
locus of failure, such as identifying a specific node $v\in V$ within
the workflow topology $\mathcal{G}_{t}$ that is causing systemic
data extraction errors. The semantic gradient is formulated as: 
\begin{equation}
g_{\text{sem}}=\mathcal{T}_{\text{critic}}\left(\Phi,\text{Tr}_{t},\hat{R}\left(\theta_{t}\right)\right).
\end{equation}

\subsubsection{Candidate Generation (The Optimiser Step)}

The teacher then synthesises a candidate plan conditioned on $g_{\text{sem}}$.
Acceptance is handled by the hill-climbing rule of Section \ref{subsec:Greedy-Hill-Climbing-and-Tabu}.
The candidate plan is given by 
\begin{equation}
\theta_{t+1}=\mathcal{T}_{\text{optimiser}}\left(\Phi,g_{\text{sem}},\theta_{t}\right).
\end{equation}

\subsubsection{Greedy Hill-Climbing}\label{subsec:Greedy-Hill-Climbing-and-Tabu}

The optimiser maintains the best-so-far plan $\theta_{\text{best}}$
and accepts a candidate $\tilde{\theta}$ only if $\hat{R}(\tilde{\theta})<\hat{R}(\theta_{\text{best}})$.
Empirically the teacher's strong prior on $\Theta_{\mathcal{T}}$
ensures rapid convergence without revisiting failed configurations.
The framework extends to Tabu constraints if needed for weaker teachers.

\subsection{Theoretical Guarantees: Resource-Bounded PAC Learning}

Because the space of all possible Python scripts and prompts is infinite,
we establish theoretical convergence guarantees by modelling the teacher’s
optimisation loop within a bounded PAC learning framework.

\subsubsection{Assumption 1 - Bounded Turing Computability and Realisability}\label{subsec:Assumption-1}

We assume the target objective admits an optimal plan $\theta^{*}$
of bounded description length (in context tokens and DAG depth), yielding
a finite hypothesis class $\Theta_{L}$ of valid plans whose nodes
are each labelled either as an LLM call (governed by $p\in\mathcal{P}$)
or a code call (governed by $c\in\mathcal{C}$). Fan-out/fan-in consensus
subgraphs are expressible without additional node types. We further
assume realisability: some $\theta^{*}\in\Theta_{L}$ satisfies $\hat{R}(\theta^{*})=0$.

\subsubsection{Lemma 1 - Occam's Razor Bound for Finite Hypothesis Classes}\label{subsec:Lemma-1}

We restate the classical realisable-case Occam bound \cite{blumer1987occam}
for reference, as it underpins the sample-complexity analysis that
follows. Under Assumption \ref{subsec:Assumption-1}, to ensure $R(\theta)\leq\epsilon$
with confidence $1-\delta$,
\begin{equation}
m\geq\tfrac{1}{\epsilon}\left(\ln|\Theta_{L}|+\ln\tfrac{1}{\delta}\right).\label{eq:sample-complexity}
\end{equation}
 The utility of this bound depends on a tight characterisation of
$|\Theta_{L}|$, which Corollary \ref{subsec:Corollary-1} supplies.

\subsubsection{Corollary 1 - The LLM Prior as an Offloading Feasibility Map}\label{subsec:Corollary-1}

The contribution of this subsection is a mechanistic characterisation
of the effective hypothesis space $\Theta_{\mathcal{T}}$ searched
by the teacher under the capability offloading principle, which when
substituted into Lemma \ref{subsec:Lemma-1} yields the sample-complexity
bound observed empirically in Section \ref{subsec:Experiment-2:-Sample}.

Although $\left|\Theta_{L}\right|$ is astronomically large due to
the combinatorial space of executable code and natural-language prompts,
the frontier teacher $\mathcal{T}$ does not search $\Theta_{L}$
uniformly. Under the capability offloading principle (Section \ref{subsec:The-Capability-Offloading}),
the teacher's pre-training knowledge functions specifically as a learned
map from subtask descriptions to offloading feasibility: for each
candidate node $v$, the teacher estimates whether $v$ is reliably
executable by deterministic code and selects the partition accordingly.

The cardinality of $\Theta_{\mathcal{T}}$ is therefore bounded not
by the prompt or code space, but by the number of distinct subtask
types in the domain. If the domain admits $k$ such types (e.g., operand
extraction, arithmetic, unit conversion, formatting), the per-type
choice is ternary (code, LLM call, or consensus subgraph), yielding
$|\Theta_{\mathcal{T}}|=\text{O}(3^{k})$. Granularity operations
reshape the partition but do not enlarge the per-type choice set.
Substituting into Lemma 1, 
\begin{equation}
m\geq\tfrac{1}{\epsilon}\left(k\ln3+\ln\tfrac{1}{\delta}\right),
\end{equation}
 linear in $k$ rather than logarithmic in the raw prompt or code
spaces. At $\epsilon=\delta=0.1$ this evaluates to $m\gtrsim11k+23$.
The empirical convergence at $m=3$ is therefore faster than the worst-case
guarantee, consistent with the teacher's strong prior on $\Theta_{\mathcal{T}}$.

\section{Experiment Evaluation}\label{sec:Experiment-Evaluation}

\subsection{Experimental Setup}

We implement the agentic workflows on a graph-based orchestrator that
exposes $\mathcal{G}$, $\mathcal{P}$, and $\mathcal{C}$ to programmatic
manipulation.

\subsubsection{Models}

The student is a heavily quantised Qwen-2.5-1.5B, simulating on-premises
enterprise deployment under strict data-governance constraints. The
teacher is Kimi 2.5, accessed via API during the offline optimisation
phase only. Its long-context reasoning is required to ingest multi-step
execution traces and emit syntactically valid JSON workflow topologies.
Each configuration is evaluated over $n=3$ independent runs with
different random seeds. Error bars show min, max, and mean $\pm$
standard deviation.

\subsubsection{Datasets}\label{subsec:Datasets}

We use GSM-Hard as a controlled proxy for structured-reasoning tasks
(mirroring the deterministic calculation requirements of financial-statement
extraction and compliance verification). To target the student's failure
modes, we (i) isolate questions the unoptimised student fails zero-shot,
(ii) randomly select a seed question and use the teacher to synthesise
structurally and mathematically similar questions, and (iii) partition
the synthesised set into disjoint training and held-out test sets;
only the held-out set is used for the reported accuracies.

\subsection{Experiment 1: Architectural Superiority}

The first experiment evaluates the performance of dynamic topological
optimisation against static prompt optimisation baselines. We compared
the baseline zero-shot student agent, a state-of-the-art prompt optimiser
(DSPy), and our proposed SGDe framework. Accuracy was measured as
the exact match of the final numerical output over the test data.

\begin{figure}
\begin{centering}
\includegraphics[width=0.85\columnwidth]{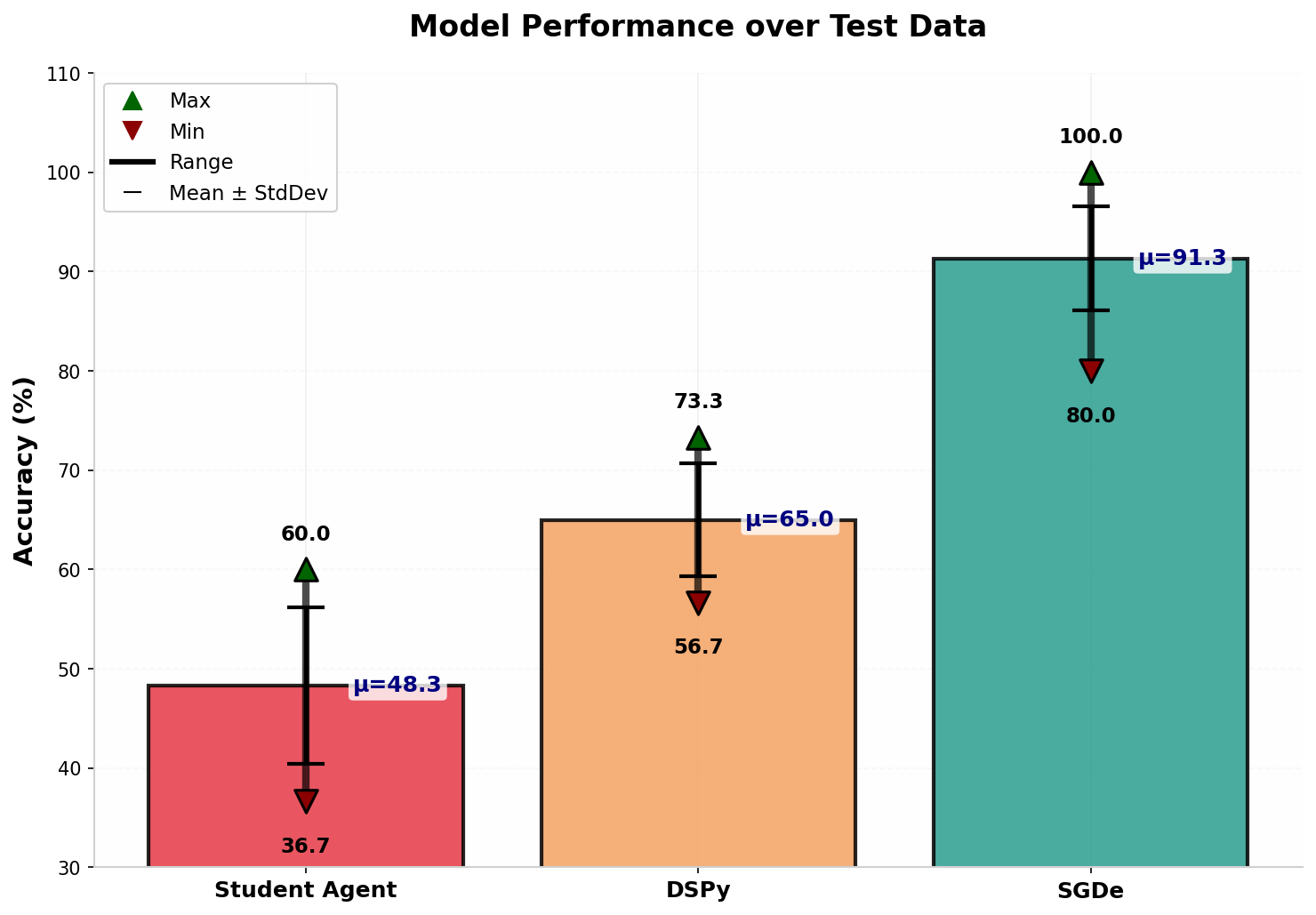}
\par\end{centering}
\caption{Accuracy comparison on the challenging GSM-Hard derived test set at
$m=5$ (see Section \ref{subsec:Datasets} for details; $m=3$ results
appear in Figure \ref{fig:sample-size}). The dynamic workflow topology
of the SGDe framework ($\mu=91.3\%$) significantly outperforms the
static architectures of both the DSPy prompt optimiser ($\mu=65.0\%$)
and the baseline student agent ($\mu=48.3\%$). }\label{fig:student-dspy-sgm}
\end{figure}

Figure \ref{fig:student-dspy-sgm} reports the result. The baseline
student reaches $\mu=48.3\%$ (range 36.7-60.0\%). DSPy lifts this
to $\mu=65.0\%$ but plateaus, limited by the SLM's native arithmetic.
SGDe reaches $\mu=91.3\%$ at $m=5$ (range 80.0-100.0\%), a +26.3\%
absolute improvement, by offloading arithmetic to deterministic code
nodes $\mathcal{C}_{t}$ and wrapping variance-sensitive reasoning
steps in fan-out/fan-in consensus subgraphs. At $m=3$ (Section \ref{subsec:Experiment-2:-Sample})
SGDe reaches $\mu=99.3\%$ (+34.3\%). Across all winning harnesses,
the teacher combined offloading with consensus. It never produced
a winner using either mechanism in isolation, confirming that the
two address non-overlapping failure modes.

\subsection{Experiment 2: Sample Complexity and PAC Learning Bounds}\label{subsec:Experiment-2:-Sample}

We sweep training sample sizes $m\in\{3,5,10\}$ to test Corollary
\ref{subsec:Corollary-1}. At $\epsilon=\delta=0.1$ and $k\approx3$
to $5$, the bound yields $m\gtrsim11k+23$. The small-$m$ empirical
regime is therefore faster than the worst-case guarantee, consistent
with the teacher's prior on $\Theta_{\mathcal{T}}$ rather than predicted
by the bound.

\begin{figure}
\begin{centering}
\includegraphics[width=0.85\columnwidth]{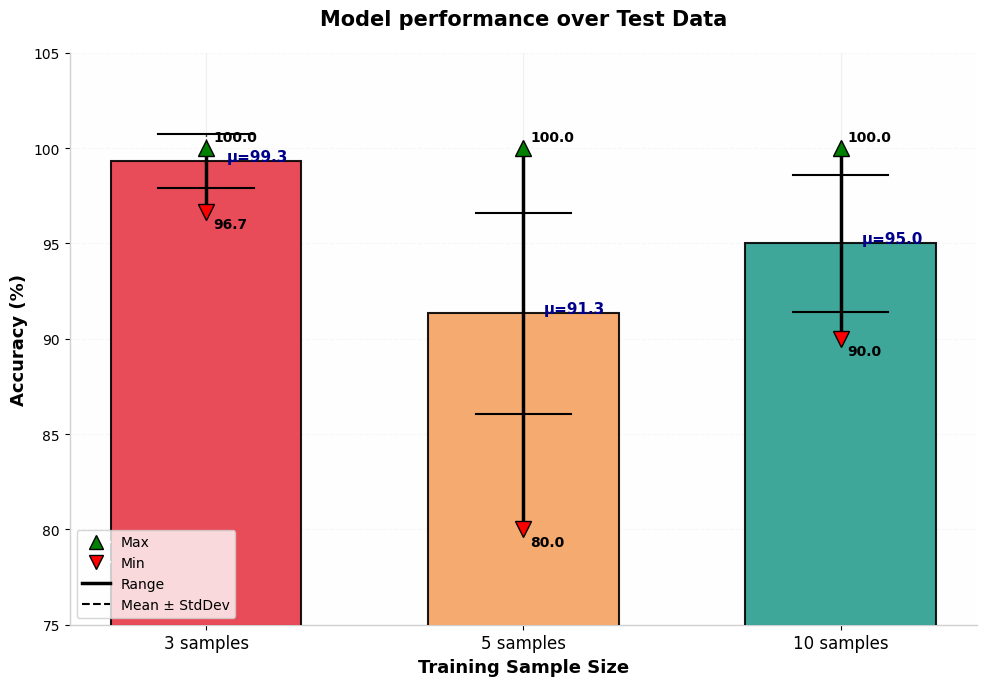}
\par\end{centering}
\caption{Impact of training sample size on test accuracy. The SGDe framework
demonstrates high data efficiency, peaking at a mean accuracy of 99.3\%
with only 3 training samples, empirically validating the proposed
PAC learning bounds for discrete workflow optimisation. }\label{fig:sample-size}
\end{figure}

Figure \ref{fig:sample-size} reports the result. SGDe peaks at $\mu=99.3\%$
at $m=3$, dips to $\mu=91.3\%$ with higher variance at $m=5$, and
stabilises at $\mu=95.0\%$ at $m=10$. The non-monotonicity reflects
batch composition mattering more than batch size in the small-sample
regime. Even the $m=5$ worst case (80.0\%) still exceeds the DSPy
maximum (73.3\%).

\subsection{Experiment 3: Boundary Conditions}\label{subsec:Boundary-Conditions}

\begin{figure}
\begin{centering}
\includegraphics[width=0.85\columnwidth]{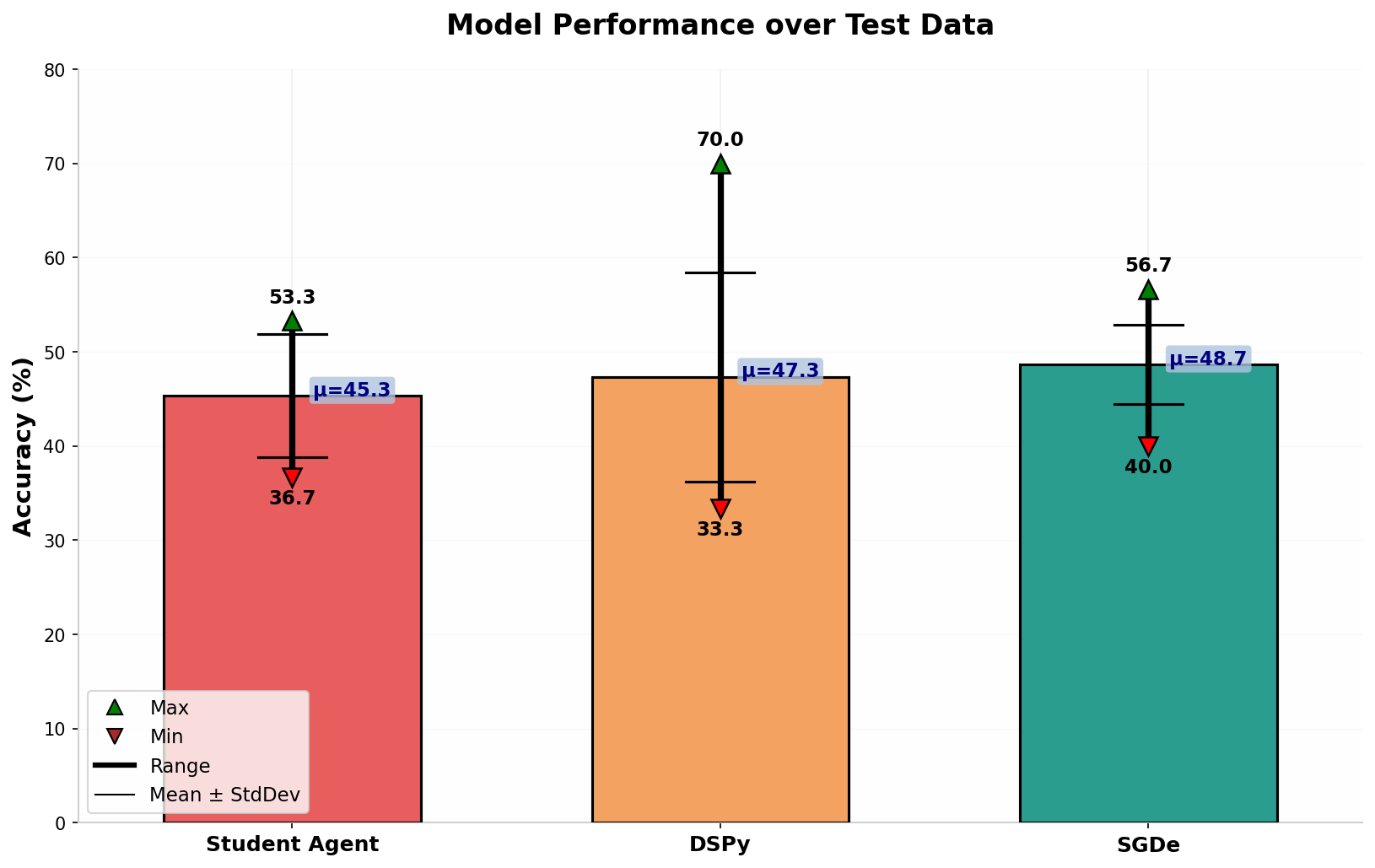}
\par\end{centering}
\caption{Validation of boundary conditions on a mixed-task test set ($Q_{1}$
vs $Q_{2}$). Both SGDe ($\mu=48.7\%$) and DSPy ($\mu=47.3\%$) fall
to near-baseline accuracy when the task family is structurally heterogeneous,
empirically confirming the realisability-boundary prediction of Corollary
1 and motivating the mixture-of-topologies extension.}\label{fig:seed=00003D2}
\end{figure}

Corollary 1 predicts that as the task family becomes structurally
heterogeneous, the number of distinct subtask types $k$ grows and
the teacher's prior on $\Theta_{\mathcal{T}}$ fragments. To validate
this prediction empirically, we evaluate on a mixed-task test set
combining synthesised questions from two structurally divergent seeds
($Q_{1}$ and $Q_{2}$). Figure \ref{fig:seed=00003D2} shows the
result. The baseline student reaches $\mu=45.3\%$, DSPy reaches $\mu=47.3\%$
(range 33.3-70.0\%), and SGDe reaches $\mu=48.7\%$ (range 40.0-56.7\%).
Both methods fall to near-baseline accuracy, confirming that no single
compiled DAG handles divergent task families well. The result empirically
motivates the mixture-of-topologies extension noted in Section \ref{sec:Conclusion}:
routing each query to a domain-specialised SGDe subgraph keeps $k$
small within each subgraph and preserves the small-$m$ convergence
regime of homogeneous tasks.

\section{Conclusion}\label{sec:Conclusion}

We introduced Semantic Gradient Descent (SGDe), a teacher-student
framework that compiles the agent harness $\theta=\{\mathcal{G},\mathcal{P},\mathcal{C}\}$
offline. At each node, the teacher selects adaptively among prompt
refinement, capability offloading, and structural consensus. The execution
plan thus becomes a partition of the task between the probabilistic
student and the deterministic runtime. The PAC analysis bounds $|\Theta_{\mathcal{T}}|=\text{O}(3^{k})$,
explaining convergence from as few as three training examples and
predicting graceful degradation as $k$ grows. Empirically, SGDe reaches
91.3\% at $m=5$ and 99.3\% at $m=3$ on the GSM-Hard-derived test
set, a +26.3\% to +34.3\% absolute improvement over DSPy, with every
winning harness combining offloaded code nodes and consensus subgraphs.
At deployment, the compiled harness executes three student calls per
query in one parallel DAG layer and zero frontier-API calls, so the
system is frontier-free at runtime. Training requires only one successful
teacher call (configured upper bound nine). Empirical validation is
restricted to GSM-Hard-style structured-reasoning tasks. The enterprise
scenarios above are motivation rather than validated use cases. Extension
to heterogeneous workflows via a mixture-of-topologies architecture
\cite{wang2024mixtureofagentsenhanceslargelanguage} remains future
work.

\bibliographystyle{ieeetr}
\bibliography{bibliography}

\end{document}